\documentclass[10pt,twocolumn,letterpaper]{article}

\usepackage[pagenumbers]{cvpr} % To force page numbers, e.g. for an arXiv version

\definecolor{cvprblue}{rgb}{0.21,0.49,0.74}
\usepackage[pagebackref,breaklinks,colorlinks,allcolors=cvprblue]{hyperref}

\def\paperID{*****} % *** Enter the Paper ID here
\def\confName{CVPR}
\def\confYear{2026}

\newcommand{\name}{Banana100}

\usepackage{cuted} % for teaser figure

\usepackage{tcolorbox}
\usepackage{xcolor}

\definecolor{PhotoshopUIGray}{HTML}{535353}   % background

\newtcolorbox{promptbox}[1][]{
  colback=gray!5,        % Light gray background
  colframe=PhotoshopUIGray,     % Photoshop grey color
  coltitle=white,        % Title color
  fonttitle=\bfseries,   % Title font style
  boxrule=0.5mm,         % Border width
  arc=2mm,               % Rounded corners radius
  title=#1,              % Allow manual title (optional)
  fontupper=\mdseries\scriptsize\setlength{\parskip}{0.2em}, 
  #1                     % Allow extra options
}

\title{Banana100: Breaking NR-IQA Metrics by 100 Iterative Image Replications with\\Nano Banana Pro}

\author{Kenan Tang, Praveen Arunshankar, Andong Hua, Anthony Yang, Yao Qin\\
University of California, Santa Barbara\\
{\tt\small kenantang@ucsb.edu, yaoqin@ucsb.edu}
}

\begin{document}
\maketitle

\begin{strip}
\centering
    {\includegraphics[width=1\textwidth]{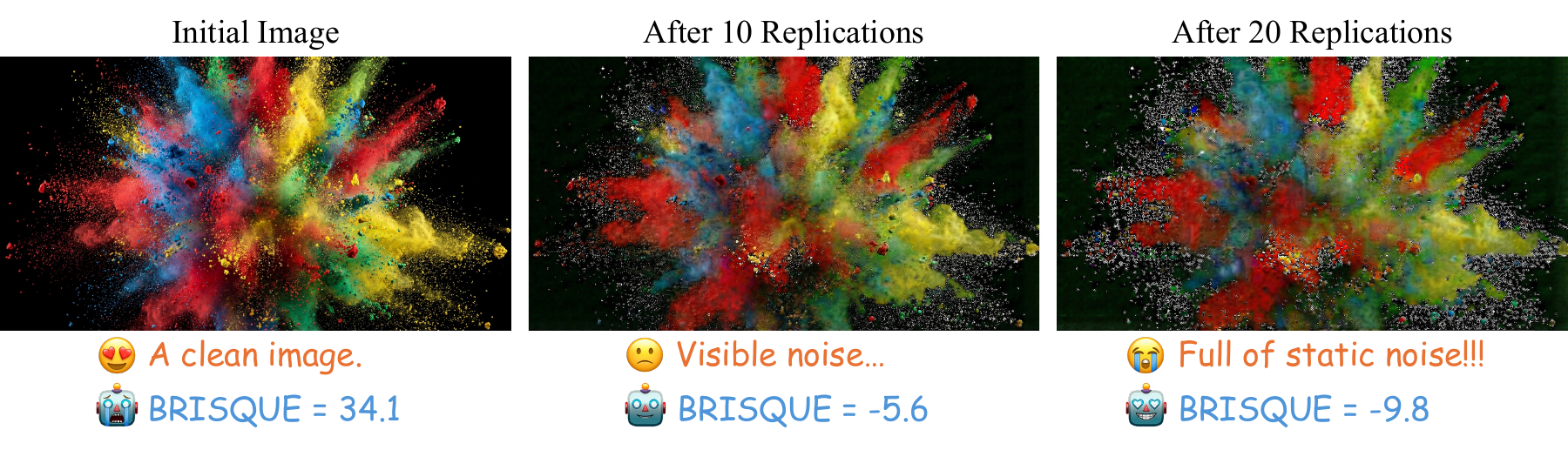}}
\captionof{figure}{\textbf{Iteratively replicating an image using Nano Banana Pro severely degrades an image, but BRISQUE assigns better (lower) scores to images of worse quality.} BRISQUE is a No-Reference Image Quality Assessment (NR-IQA) metric that has been widely used to assess the quality of AI-generated images. This counter-intuitive failure is pervasive across diverse NR-IQA metrics and image textures.}
\label{fig:degradation}
\end{strip}

\begin{abstract}
The multi-step, iterative image editing capabilities of multi-modal agentic systems have transformed digital content creation. Although latest image editing models faithfully follow instructions and generate high-quality images in single-turn edits, we identify a critical weakness in multi-turn editing, which is the iterative degradation of image quality. As images are repeatedly edited, minor artifacts accumulate, rapidly leading to a severe accumulation of visible noise and a failure to follow simple editing instructions. To systematically study these failures, we introduce \name, a comprehensive dataset of 28,000 degraded images generated through 100 iterative editing steps, including diverse textures and image content. Alarmingly, image quality evaluators fail to detect the degradation. Among 21 popular no-reference image quality assessment (NR-IQA) metrics, none of them consistently assign lower scores to heavily degraded images than to clean ones. The dual failures of generators and evaluators may threaten the stability of future model training and the safety of deployed agentic systems, if the low-quality synthetic data generated by multi-turn edits escape quality filters. We release the full code and data to facilitate the development of more robust models, helping to mitigate the fragility of multi-modal agentic systems.\footnote{\url{https://huggingface.co/datasets/kenantang/Banana100}}
\end{abstract}

\section{Introduction}

AI-based image-text-to-image (IT2T) models has transformed digital content creation~\cite{bfl2025flux2, wu2025qwen, seedream2025seedream, cao2025hunyuanimage, liu2025magicquill, liu2025magicquillv2}. These tools allow users to both create new images and iteratively refine them, promising a high degree of creative freedom. This multi-step editing paradigm is further facilitated by the rise of multi-modal agentic systems~\cite{ye2026agent, lin2025jarvisart, zuo2025kagent, yao2026photoagent}, where autonomous systems composed of a generator (an image editing model) and an evaluator (an image quality assessor) can orchestrate complex image refinement processes.
 
While modern models such as Nano Banana Pro~\cite{google2025nanobananapro} demonstrate impressive image quality in single-turn edits, we identify a critical and underexplored failure mode in the multi-turn scenario, which is \textbf{iterative degradation}. During each editing pass, image generators always introduce minor, often imperceptible artifacts~\cite{almog2025reed, liao2025freqedit}. When an output image is fed back into the model again for subsequent edits, these artifacts accumulate into visible quality degradation, such as static noise (\Cref{fig:degradation}), greenish tint (\Cref{fig:failure_modes}), or scatter points (\Cref{fig:other_models}). Our experiments reveal that after around 5 to 10 steps, Nano Banana Pro quickly starts to suffer from the following two failures:
 
\begin{enumerate}
    \item \textbf{Visual Quality Degradation:} High-frequency details are distorted, and visual artifacts emerge in regions that were never targeted for editing.
    \item \textbf{Instruction Following Failure:} The model's capacity to faithfully execute editing prompts progressively deteriorates, failing to follow even very simple prompts, such as adding an apple on a table (\Cref{fig:failure_modes}).
\end{enumerate}
 
Of greater concern, methods that could potentially serve as the the evaluator component in agentic pipelines prove unreliable for detecting these failure patterns. Out of the 23 popular no-reference image quality assessment metrics (NR-IQA) we examined (\Cref{sec:nr_iqa_methods_fail_to_quantify_degradation}), only 2 consistently detected the degradation. Other 21 metrics reported higher quality for noisy images than clean images. As an alarming example, simply replicating an initial image can reach a better (lower) BRISQUE score, despite introducing severe noise and corrupting the original image content (\Cref{fig:degradation}). While the clean initial image received a BRISQUE score of 34.1, the noisy image after 20 replications received a far lower (better) BRISQUE score of -9.8. The scores are completely flipped compared to human-perceived image quality.

The failures of both the generator and the evaluator allow the degradation to silently leak into datasets without being detected. As an example, the multi-step subset of Pico-Banana-400K~\cite{qian2025picobanana400klargescaledatasettextguided} exhibited obvious distortions of object textures and human faces, especially after five~\cite{apple2026picobanana10006turn5} or six~\cite{apple2026picobanana10006turn6} editing steps.\footnote{The references point to only two example images, but in this dataset, many other images after 5 steps generally suffer from similar degradation.} The potential negative consequences are profound. In particular, we highlight two possible downstream effects. First, on the training side, as AI-edited content proliferates, the future training data may become increasingly noisy. If evaluators fail to filtering out noisy data, model collapse could be accelerated in subsequent image generation models~\cite{shumailov2024ai, yoon2024model}. Second, on the inference side, agentic systems are known to be fragile over a long horizon~\cite{shao2025your, chen2026swe}. If the degraded images escape the quality checks, the fragility could be further exacerbated.

To address these challenges, our three contributions are:
 
\begin{enumerate}
    \item \textbf{Large-scale dataset of iterative degradation:} We introduce \name, a dataset constructed by iteratively editing 13 diverse initial images using 100 editing steps with various instructions, yielding 28,000 images at a cost of \$4,000 (\Cref{sec:the_dataset}). Other than Nano Banana Pro, we also confirmed the generalizability of the dataset construction pipeline to more IT2T models (\Cref{sec:other_image_editing_models_fail_similarly}).
    \item \textbf{Systematic failure mode taxonomy:} With diverse initial images, \name\ demonstrates multi-step visual quality degradations and instruction-following failure modes, which we systematically categorize into sub-object, object, and image levels (\Cref{sec:analysis_of_instruction_following_failures}). 
    \item \textbf{Identification of flawed NR-IQA metrics:} Beyond generator failure, \name\ helps to quantitatively identify existing NR-IQA metrics that assign counter-factually good scores for low-quality images (\Cref{sec:noise_quantification_and_NR_IQA_failures}). This will help researchers avoid falsely reporting an improvement in image quality when the metrics are actually confounded by model-induced degradation, facilitating the development of more robust NR-IQA metrics.
\end{enumerate}

\section{The Banana100 Dataset}
\label{sec:the_dataset}

We constructed \name\ by iteratively editing images using Nano Banana Pro. Each initial image was edited by a prompt, and then the output served as the input for the next editing step. Each run consists of 100 editing steps.

\subsection{Initial Images}

We collected a set of high quality initial images with the following 5 requirements. First, the initial images should be in high resolution, with minimal compression artifacts to start with. Second, the initial images should be free from potential copy-right violations. Third, the initial images themselves should be AI-generated, aligning with the realistic scenario that a user first generates an image with text prompts and then edits these images multiple times with additional instructions. Fourth, the images should cover a diverse range of topics and textures, stress-testing the model's capability in exact replication. Finally, we deliberately excluded photorealistic faces of humans, as the distortions on real faces are usually visually unpleasant and disturbing~\cite{apple2026picobanana10006turn6}.

Following these requirements, we curated 13 initial images, all in at least 2K resolution (\Cref{tab:initial_images}). 11 were generated by Nano Banana Pro, with manually refined prompts to cover diverse topics and textures. 2 were generated using SPICE~\cite{tang2025spice}, a method that excels at generating highly-resolution and factually correct anime-style images.

\begin{table*}
  \scriptsize
  \caption{\textbf{The initial images cover diverse content and challenges.} The top part of the table includes 11 photorealistic images generated by Nano Banana Pro, and the bottom part includes 2 animation-style images generated by SPICE. The resolutions are width$\times$height.}
  \label{tab:initial_images}
  \centering
  \begin{tabular}{@{}llll@{}}
    \toprule
    \textbf{Name} & \textbf{Image Content} & \textbf{Challenges} & \textbf{Resolution} \\
    \midrule
    Building & A skyscraper & Preservation of highly-regular grid patterns and aerial perspective & 3392$\times$5056 \\
    Dongpo & A plate of Chinese potstickers & Preservation of multi-scale food structure and texture & 5504$\times$3072 \\
    Ekphrasis & A still life painting & Preservation of diverse textures of the same type of object & 5632$\times$3072 \\
    Fog & A misty forest & Preservation of texture details under lowered color contrast by haze & 5504$\times$3072 \\
    Holi & Exploding colorful Holi powder & Preservation of high color contrast and particle textures & 5632$\times$3072 \\
    Library & Interior of a library & Preservation of deep shadows and shafts of light & 5632$\times$3072 \\
    Moss & Tree bark covered in moss and lichen & Preservation of soft and non-periodic texture details & 4800$\times$3584 \\
    Peacock & A peacock feather& Preservation of iridescent texture details & 4800$\times$3584 \\
    Rice & Rice terraces during sunset & Preservation of reflections and repeated patterns with variations & 5504$\times$3072 \\
    Sand & A sand dune at twilight & Preservation of smooth color gradients & 5504$\times$3072 \\
    Table & An empty wooden table & Addition of diverse objects while preserving the background & 5504$\times$3072 \\
    \midrule
    Kokoro & A standing animation character & Preservation of asymmetric design and clean stylistic colors & 1664$\times$2432 \\
    Yuiman & A grid of 9 diverse headshot poses & Preservation of 4-colored gradients in the eyes and the grid layout & 3000$\times$3000 \\
    \bottomrule
  \end{tabular}
\end{table*}

Note that the conclusions drawn from a deliberately curated set of AI-generated initial images may not directly generalize if the initial images were real-life image with potential compression artifacts, due to the known gap between the two distributions~\cite{adamkiewicz2026pretty}. We leave to future work the exploration of using real-life images for the initial images.

\subsection{Iterative Editing Prompts}
\label{sec:iterative-editing-prompts}

We designed the iterative editing prompts to test the preservation of image quality and the evaluation of it with minimal confounders. One great confounder for the NR-IQA metrics turns out to be the image content. While the initial images are all free from visible noise, some quality metrics provide dramatically different scores for these images. As an example, among all 13 initial images, the Yuiman image has a lowest BRISQUE score of -3.18, while Kokoro has a highest BRISQUE score of 41.1. However, both images were generated with SPICE, and no visible noise is present. 

Therefore, to minimize the confounding effects of image content on the quality, we primarily conducted the \textbf{replication} runs, where the model was asked to ``Produce an exact replica of the provided image, with no alterations.'' This focus on a seemingly simple replication task is justified by our pilot study, which revealed that replication leads to noise patterns qualitatively similar to the ones observed with prompts that actually change the semantic content of an image, such as adding objects.

Besides this straightforward prompt with the default hyperparameter set, we also investigated 5 more variants:

First, we changed the phrasing of the replication prompt. While the straightforward prompt quantitatively reproduced the failure patterns aligning with general user experience, we would like to test for the sensitivity of vision-language models to the prompt phrasing~\cite{mo2024dynamic, hua-etal-2025-flaw}.

Second, we also included multi-step replication operations that transform an image back to its original content using more than one step. For example, horizontally mirroring an image twice ends up with the original image. This variant was motivated by the observation that when the model is asked to explicitly change one region on the image, the changed region will suffer less from degradation (\Cref{sec:object_level_failure_modes}). Hence, explicitly asking the model to edit the full image might help mitigating the noise accumulation.

Third, we further relaxed the requirements on replication by including multi-step reconstruction prompts. These methods are popular in the user community for their potential in denoising a model-edited noisy image. For example, the model is asked to extract simplified color patches in a first step and to extract edge information in a second step. Then, in the third step, the model is asked to reconstruct a photorealistic image from the color patches and the edges. We observed that this method empirically resulted in noise-free images, but the image content was hardly preserved over multiple iterations. Since these methods do not align with the fundamental user requirements of preserving both the quality and the semantic content, we only included a limited number of such runs in the dataset as a reference, but we did not use these runs for image quality assessment.

Fourth, we tested with alternative values of three hyperparameters in the Nano Banana Pro model, including seed, temperature, and resolution. For the seed, either a fixed seed was used throughout the editing steps, or a different seed was provided for each step. This was motivated by observations in our pilot studies that certain images and methods suffer from artifacts when a fixed seed was used throughout the steps, although these artifacts cannot be reliably reproduced due to the black-box nature of proprietary models. The temperature was either set at 0 or 0.4. The resolution was set to be one of the three options allowed by the API, including 1K, 2K, and 4K. The resolution could only be chosen from these three strings, instead of specified as numeric values. The majority of the dataset was generated with the default resolution of 2K. We used alternative resolutions or interleaved different resolutions (switching periodically in the order of 1K, 2K, and 4K for each step) for a small number of runs, only to investigate the impact of resolution.

Finally, to better align with the real use cases while keeping confounders minimal, we also used prompts that change only a small region on an image. The Table image was chosen for two tasks of adding the same type of fruit (\textbf{add-apples} run) or adding different fruits (\textbf{add-100-fruits} run).

All settings above were run with 100 steps, each time in a separate chat session through the Nano Banana Pro API. To keep the cost from quadratically increasing, we did not include all editing steps in a same dialog session. We qualitatively discuss single-session results in \Cref{sec:image_level_failure_modes}. To ensure robust analysis, we perform 5 separate runs per setting. However, achieving a full grid search combination is costly. We primarily focused on the \textbf{replication} runs, which was available for all 12 seed images, excluding the Table image that did not include challenging textures and was thus used only for object addition (\textbf{add-apples} and \textbf{add-100-fruits}). 

Overall, the development and construction of the dataset cost over \$4,000, resulting in a dataset of 28,000 total output images. The number of images is comparable in the order of magnitude to popular IQA training and evaluation datasets, such as BID~\cite{ciancio2010no}, CLIVE~\cite{ghadiyaram2015massive}, KonIQ-10k~\cite{hosu2020koniq}, SPAQ~\cite{fang2020perceptual}, Liu13 (deblurring)~\cite{liu2013no}, Min19 (dehazing)~\cite{min2019quality}, AGIQA-3K (image generation)~\cite{li2023agiqa}, and UHD-IQA~\cite{hosu2024uhd}. Our dataset is smaller than some of the existing IQA datasets, such as SRIQA-Bench (super-resolution)~\cite{chen2025toward}, KADIS-700K~\cite{lin2019kadid}, and AVA~\cite{murray2012ava}. However, the high image resolution in our dataset allows the extraction of multiple patches from each image for training or evaluation~\cite{goring2023quality}, further increasing the effective size of our dataset. 

\subsection{Model Selection}

We selected Nano Banana Pro for its high popularity and its high rank on the Image Edit Arena~\cite{arena2026imageedit}. While Nano Banana Pro was our primary focus for dataset development, we also tested its successor, Nano Banana 2~\cite{google2026nanobanana2}, together with other open-source models at a smaller scale to demonstrate their qualitative similarities and differences (\Cref{sec:other_image_editing_models_fail_similarly}). 

We leave the investigation of other agentic image editing systems~\cite{ye2026agent, lin2025jarvisart, zuo2025kagent, yao2026photoagent} as future work. However, our focus on the underlying image-editing model deployed in those systems should shed light on the expected degradation behavior of agentic image editing systems. Notably, the evaluation of some agentic image editing systems~\cite{zuo2025kagent, yao2026photoagent} heavily relied on the NR-IQA metrics such as BRISQUE and NIQE, which we reveal as deeply flawed (\Cref{sec:nr_iqa_methods_fail_to_quantify_degradation}). 

Our dataset is complementary to the existing large scale datasets derived from Nano Banana~\cite{qian2025picobanana400klargescaledatasettextguided} and Nano Banana Pro~\cite{zuo2025nano, wei2025mico}. Instead of curating a dataset for the utility of high quality images, we highlight the controlled quality degradation that is unique to our dataset.

\subsection{Reasoning Summary}

\begin{figure}[h]
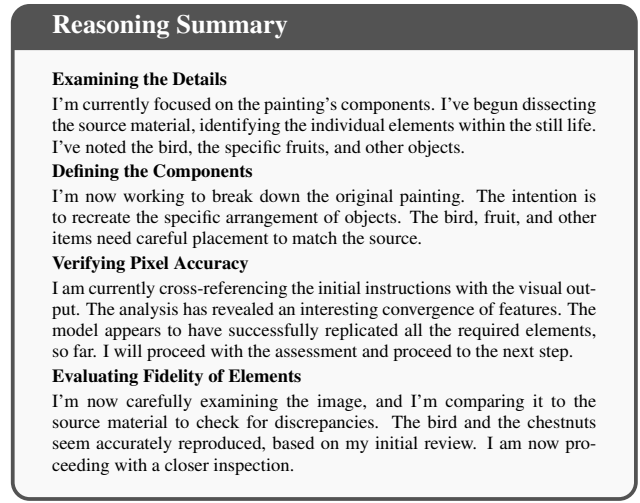

\centering
\begin{promptbox}[title=Reasoning Summary]
\textbf{Examining the Details}

I'm currently focused on the painting's components. I've begun dissecting the source material, identifying the individual elements within the still life. I've noted the bird, the specific fruits, and other objects.

\textbf{Defining the Components}

I'm now working to break down the original painting. The intention is to recreate the specific arrangement of objects. The bird, fruit, and other items need careful placement to match the source.

\textbf{Verifying Pixel Accuracy}

I am currently cross-referencing the initial instructions with the visual output. The analysis has revealed an interesting convergence of features. The model appears to have successfully replicated all the required elements, so far. I will proceed with the assessment and proceed to the next step.

\textbf{Evaluating Fidelity of Elements}

I'm now carefully examining the image, and I'm comparing it to the source material to check for discrepancies. The bird and the chestnuts seem accurately reproduced, based on my initial review. I am now proceeding with a closer inspection.
\end{promptbox}
\caption{\textbf{The reasoning summary from Nano Banana Pro appears as clear-cut generation and evaluation sections.} The bold text are section titles, copied verbatim from the reasoning summary from the Nano Banana Pro API. In this example, the first two sections are dedicated to image generation, whereas the last two sections are dedicated to the evaluation of a generated image.}
\label{fig:example_reasoning_summary}
\end{figure}

Since Nano Banana Pro is a reasoning model, a reasoning trace is generated together with the output image. As Nano Banana Pro does not reveal its full reasoning trace even in the API output, we only included the reasoning summary returned by the API in \name. The reasoning summary is broken down into multiple sections. \Cref{fig:example_reasoning_summary} shows an example, in which the final two sections perform evaluation, where the model checks if its output aligns with the prompt. In rare cases, the model mentions that the generated output does not align with the prompt and returns to a second round of generation, resulting in a larger number of reasoning summary sections. However, the more predominant pattern is that Nano Banana Pro tends to generate fully confident evaluations, even when the output image totally fails to align with the input text prompt (\Cref{sec:analysis_of_instruction_following_failures}).

\section{Analysis of Instruction Following Failures}
\label{sec:analysis_of_instruction_following_failures}

\begin{figure}
  \centering
  \includegraphics[width=1\columnwidth]{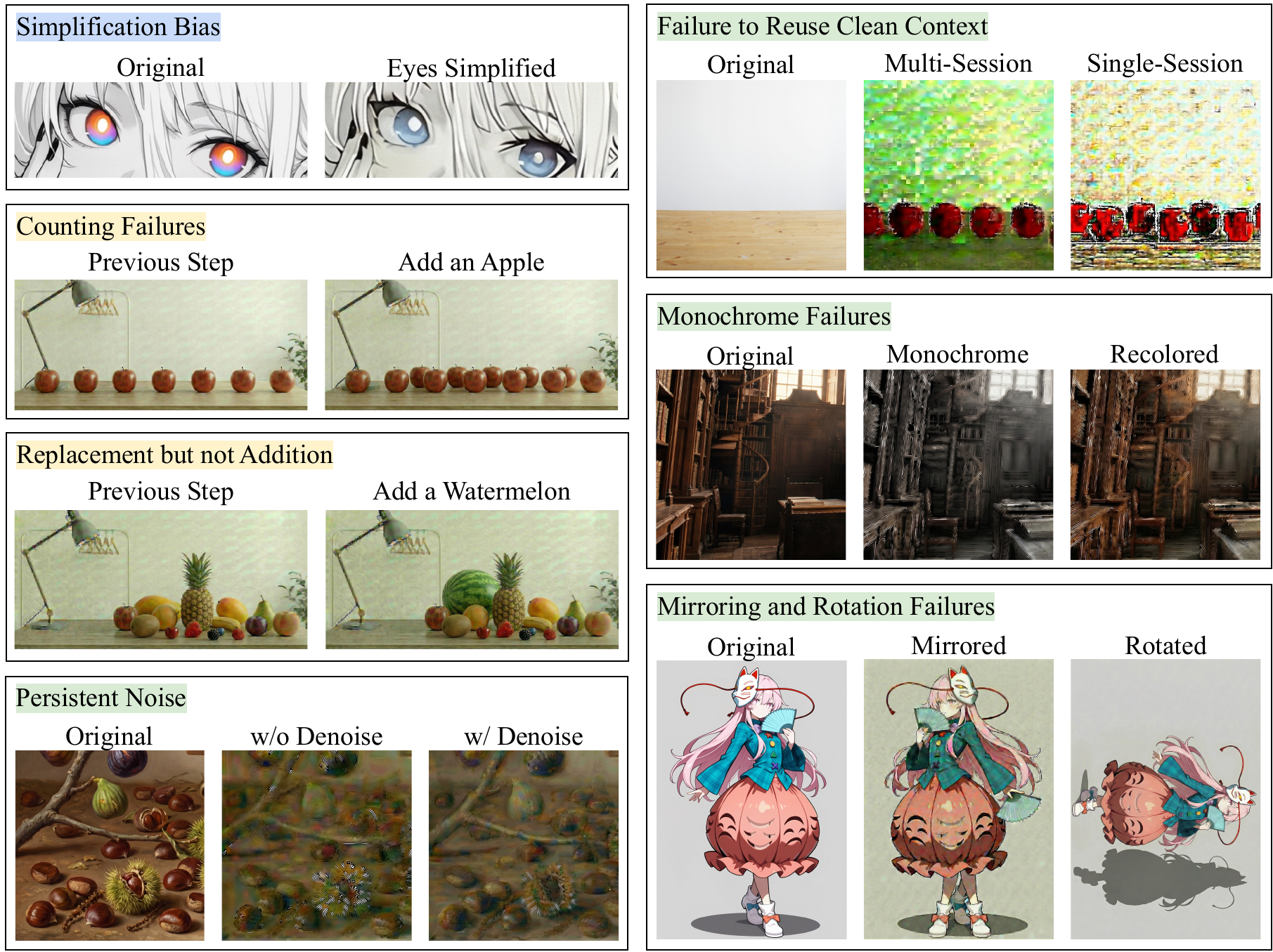} 
  \caption{\textbf{A summary of the failure modes of instruction following, categorized into sub-object level (blue), object level (yellow), and image level (green).} The images have been cropped and zoomed for visual clarity. As the failures were consistent across different runs and editing steps, we do not report the exact run index and step index for each image here. See \Cref{sec:analysis_of_instruction_following_failures} for details.}
  \label{fig:failure_modes}
\end{figure}

In this section, we qualitatively analyze the failure modes of instruction following. Other than the accumulation of global low-level noise (\Cref{fig:degradation}), Nano Banana Pro also failed to follow instructions at three different levels, dubbed as sub-object level, object level, and image level (\Cref{fig:failure_modes}). While non-exhaustive, we list the most obvious failure modes at each level and demonstrate the reasoning summary hallucinations associated to the failures. At least one example image will be provided for each failure mode, and more example images of each failure mode can be easily accessed in our publicly shared dataset.

\subsection{Sub-Object-Level Failure Modes}

In sub-object-level failures, the model failed to faithfully replicate a part of an object. This most frequently happened when a character has a complex and detailed visual design.

\paragraph{Simplification Bias.} When asked to replicate the image of a character expression grid (Yuiman), the model failed to replicate the exact eye colors after the second step. The original four eye colors (red, orange, purple, and blue) were quickly simplified to only red and blue. In the reasoning summary, we saw that the model only captured the most prominent colors (red and blue) of the eyes, ignoring the other colors (orange and purple). Interestingly, not all grids suffered from the color simplification at the same step. The color gradients on some eyes were preserved in the early steps, but all gradients eventually vanished within 5 steps.

This sub-object level failure mode reveals that maintaining character consistency remains an unresolved task. While the consistency might be improved by specifying the character details in the prompt, this approach quickly tumbles as the number of characters on an image increases.

\subsection{Object-Level Failure Modes}
\label{sec:object_level_failure_modes}

In object-level failures, the model simply failed to add an object as instructed. Two patterns are listed below.

\paragraph{Counting Failures.}

In the add-apples run, the model was asked to add an apple to the table in each step. In the early steps where the numbers of apples were as small as 7, the model already failed to add one more apple. Moreover, the evaluation section in the reasoning summary mismatched the generation failure. For 3 consecutive editing steps in one run, while the reasoning summary correctly identified 7 apples and confirmed the new total to be 8, the model did not generate a new apple. In the next editing step, the model alternatively added a full row of apple, disregarding the instruction completely.

\paragraph{Replacement but not Addition.}

In the add-100-fruits run, the model was asked to add 100 different fruits to the table, one in each step. Instead of adding the fruit, the model sometimes replaced one of the existing fruit with the new fruit, regardless of the fruit size or the relative position of the fruit (the example shows the replacement of a papaya by a watermelon in the background). The reasoning summary showed that the model did not exhaustively examine each of the existing fruits on the table. Since the full reasoning trace is not visible, we cannot confirm whether skipping some fruits during reasoning caused this replacement issue.

\paragraph{Consistent Background Degradation.} Throughout 100 editing steps, the added new object sometimes had refreshed visual quality, less affected by the worsening noise in the background. This seemed to suggest that editing an image globally might mitigate the noise accumulation and preserve the quality. This motivated us to test the roundtrip decolorization and colorization editing of an image as one of the multi-step reconstruction methods (\Cref{sec:iterative-editing-prompts}). In these edits, the model was asked to turn the image monochrome in one editing step and to color the monochrome image in a subsequent editing step, in two separate chat sessions. Although this pair of roundtrip editing steps could not preserve the original colors, this experiment setting was designed to test whether the noise can be removed and the quality can be preserved. However, the next subsection shows that this approach did not work.

These object-level failure modes reconfirm that handling spatial relationship of objects remains challenging, especially in the presence of model-induced low-level noise.

\subsection{Image-Level Failure Modes}
\label{sec:image_level_failure_modes}

In image-level failure modes, the model failed in maintaining or changing the properties defined on the whole image, such as aspect ratio or orientation. 

\paragraph{Aspect-Ratio Mismatch.} When asked to replicate the image, Nano Banana Pro almost always cropped the image in the first step. This might be due to the model requiring the side length of the output to be from a certain set of numbers. As an example, the resolution of the Ekphrasis image was changed from 5632$\times$3072 to 1408$\times$752, 2816$\times$1504, and 5632$\times$3008 for output resolutions of 1K, 2K, and 4K, respectively. The aspect ratio was changed from 0.545 to 0.534 in all 3 cases by cropping existing pixels in the input.

\paragraph{Persistent Noise.} The noise introduced over editing steps is persistent, regardless of the prompt phrasing or hyperparameter changes. Notably, explicitly including a denoising instruction in each prompt did not preserve the image quality or content over editing steps. By comparing the ``w/o Denoise'' and ``w/ Denoise'' images (both at 20 steps), we saw that both images suffer similarly from an added green tint and a loss of texture. From the reasoning summary, we saw that the model attempted denoising and removing artifacts, but it failed to denoise the output images at each step.

\paragraph{Failure to Reuse Clean Context.} One may argue that the multi-session, single-turn setting we adopted prevented the model from reusing the clean images in an earlier generation to eliminate the noise accumulated over the steps. Indeed, as it supports a large context size, the model should be able to use all past context instead of just the most recent image. However, when using a single session in the interface for the same object addition task, we saw that the generated result similarly suffered from degradation. 

\paragraph{Monochrome Failure.} When asked to make an image monochrome, the model did not convert the colors strictly to grayscale. Also, the image quality still degraded over the steps, invalidating this two-step reconstruction method.

\paragraph{Mirroring and Rotation Failures.} For multi-step replication, we chose horizontally mirroring (recovering the original image in every 2 steps) and clock-wise rotation by 90 degrees (recovering the original image in every 4 steps). The mirroring and rotation operations were performed on one realistic image (Ekphrasis) and one animation image (Kokoro). For mirroring, the model had a much lower success rate on the animation image than the realistic image. For rotation, the success rates were low for both images. For both operations, the image quality degraded similarly as with the naive replication operation. However, the reasoning summary in each step showed hallucinated confidence.

Again, all these full-image operations were motivated by their potential in preserving the image quality over editing steps. Since the obvious failures disqualified these methods from preserving image quality, we did not further quantify the exact failure rate in depth. 

\section{Noise Quantification and NR-IQA Failures}
\label{sec:noise_quantification_and_NR_IQA_failures}

Next, we focused on only the \textbf{replicate} runs for 12 initial images and attempted to use Image Quality Assessment (IQA) metrics to quantify the introduced noise. We used a subset of No-Reference IQA (NR-IQA) methods where a score can calculated based on an individual image. NR-IQA metrics requiring a reference dataset, such as FID~\cite{heusel2017gans}, were excluded. Full-Reference IQA (FR-IQA) metrics that require a pair of semantically identical images, such as PSNR~\cite{hore2010image}, LPIPS~\cite{zhang2018unreasonable}, and SSIM~\cite{wang2004image}, were also excluded. 

We note that the FR-IQA metrics could be interfered by the change of semantic content on an image (such as an addition of an object). Although we adopted a simplified setting of image replication, such interference makes FR-IQA metrics less suitable than NR-IQA metrics, when the end goal is to investigate the quality degradation regardless of the semantic content. Also, among NR-IQA metrics, the ones that are less interfered by the semantic content are more suitable for the quantification of model-induced noise (more details in \Cref{sec:two_recent_nr_iqa_methods_succeed}). 

\subsection{NR-IQA Methods Fail to Quantify Degradation}
\label{sec:nr_iqa_methods_fail_to_quantify_degradation}

\begin{table}
  \scriptsize
  \caption{\textbf{A summary of all the No-Reference Image Quality Assessment (NR-IQA) metrics we used for evaluation.} In the first part of the table, we show all the NR-IQA metrics implemented in the \texttt{pyiqa} Python library~\cite{chen2024iqapytorch}, with the only exception of MACLIP~\cite{liao2026beyond}, which is only a placeholder and raises a non-implemented error. The typical range is obtained from the \texttt{pyiqa} library, which do not necessarily correspond to the actual observed range. In the second part of the table, we include two recent NR-IQA metrics based on latest large vision-language models.}
  \label{tab:metrics}
  \centering
  \begin{tabular}{@{}lll@{}}
    \toprule
    \textbf{Metric} & \textbf{Typical Range} & \textbf{Higher is Better?} \\
    \midrule
    ARNIQA~\cite{agnolucci2024arniqa} & [0, 1] & Yes \\
    BRISQUE~\cite{mittal2012no} & [0, 150] & No \\
    CLIPIQA~\cite{wang2023exploring} & [0, 1] & Yes\\
    CNNIQA~\cite{kang2014convolutional} & [0, 1] & Yes \\
    DBCNN~\cite{zhang2018blind} & [0, 1] & Yes \\
    HyperIQA~\cite{su2020blindly} & [0, 1] & Yes \\
    ILNIQE~\cite{zhang2015feature} & [0, 100] & No \\
    LIQE~\cite{zhang2023blind} & [1, 5]& Yes \\
    MANIQA~\cite{yang2022maniqa} & [0, 1] & Yes \\
    MUSIQ~\cite{ke2021musiq}  & [0, 100] & Yes \\
    NIMA~\cite{talebi2018nima}  & [0, 10] & Yes \\
    NIQE~\cite{mittal2012making}  & [0, 100] & No \\
    NRQM~\cite{ma2017learning}  & [0, 10] & Yes \\
    PaQ-2-PiQ~\cite{ying2020patches}  & [0, 100] & Yes \\
    PI~\cite{blau20182018}  & $\geq0$ & No \\
    PIQE~\cite{venkatanath2015blind}  & [0, 100] & No \\
    Q-Align~\cite{wu2024qalign}  & [1, 5] & Yes \\
    QualiCLIP~\cite{agnolucci2024quality}  & [0, 1] & Yes \\
    TOPIQ NR~\cite{chen2024topiq}  & [0, 1] & Yes \\
    TReS~\cite{golestaneh2022no}  & [0, 100] & Yes \\
    WaDIQaM~\cite{bosse2017deep} & [-1, 0.1] & Yes \\
    \midrule
    VisualQuality-R1~\cite{wu2025visualqualityr}  & [1, 5] & Yes \\
    RALI~\cite{zhao2025reasoning}  & [1, 5] & Yes \\
    \bottomrule
  \end{tabular}
\end{table}

The NR-IQA metrics we used are summarized in \Cref{tab:metrics}. We directly used the models implemented in the \texttt{pyiqa} Python library~\cite{chen2024iqapytorch}. When multiple models trained on different datasets are available for one metric, we only used the default version as specified on the Model Card page~\cite{chen2024iqapytorch}.

\begin{figure}
    \centering
    \includegraphics[width=0.8\columnwidth]{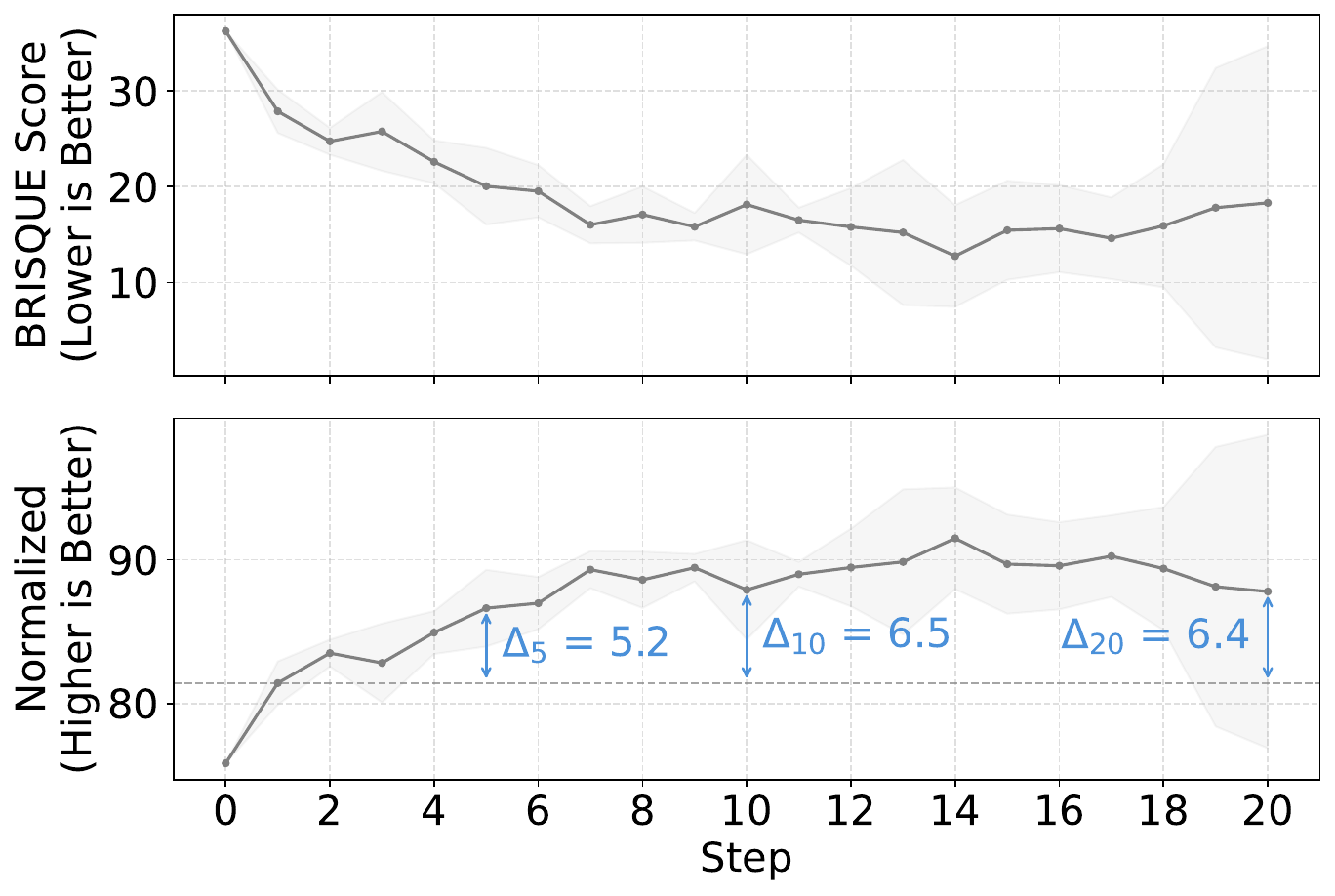}
    \caption{\textbf{We normalized NR-IQA scores (BRISQUE as an example) and calculated the difference across steps to quantify the score trend.} Please see \Cref{sec:nr_iqa_methods_fail_to_quantify_degradation} for details. The three $\Delta$ values can also be found at the intersection of the second row (Dongpo) and the second column in each heatmap of \Cref{fig:iqa_nano_banana_pro}.}
    \label{fig:delta_calculation}
\end{figure}

Since the small degradation over a single step is hard to be precisely judged by humans, we did not obtain Mean Opinion Scores (MOS) for individual images and thus did not use the Pearson Linear Correlation Coefficient (PLCC) and the Spearman Linear Correlation Coefficient (SRCC), two metrics commonly used to rank the performance of NR-IQA models. Instead, we based our evaluation on the observation that the image quality drop after multiple steps is very obvious to bare eyes (\Cref{fig:degradation}). This observation aligns with the general experience widely reported by contemporary users. Based on this observation, we define the normalized score gap $\Delta_i$ to be the normalized score of Step $i$ minus the normalized score gap of Step 1 (\Cref{fig:delta_calculation}). Here, $i$ can take values from $\{5, 10, 20\}$ but not smaller numbers, because the image quality is unambiguously decreasing for a human observer after a sufficiently large number of editing steps. The initial step was chosen to be 1 instead of 0, in order to avoid confounding effects of cropping (\Cref{sec:image_level_failure_modes}). The normalization maps the score from the typical score range to [0, 100], flipping the direction for BRISQUE, ILNIQE, NIQE, PI, and PIQE such that a higher score consistently indicates higher quality. Notably, the normalization does not change the potency of the metric in distinguishing image quality, but it only provides a consistent score scale and direction for the convenience of comparison.

\begin{figure}
  \centering
  \includegraphics[width=\columnwidth]{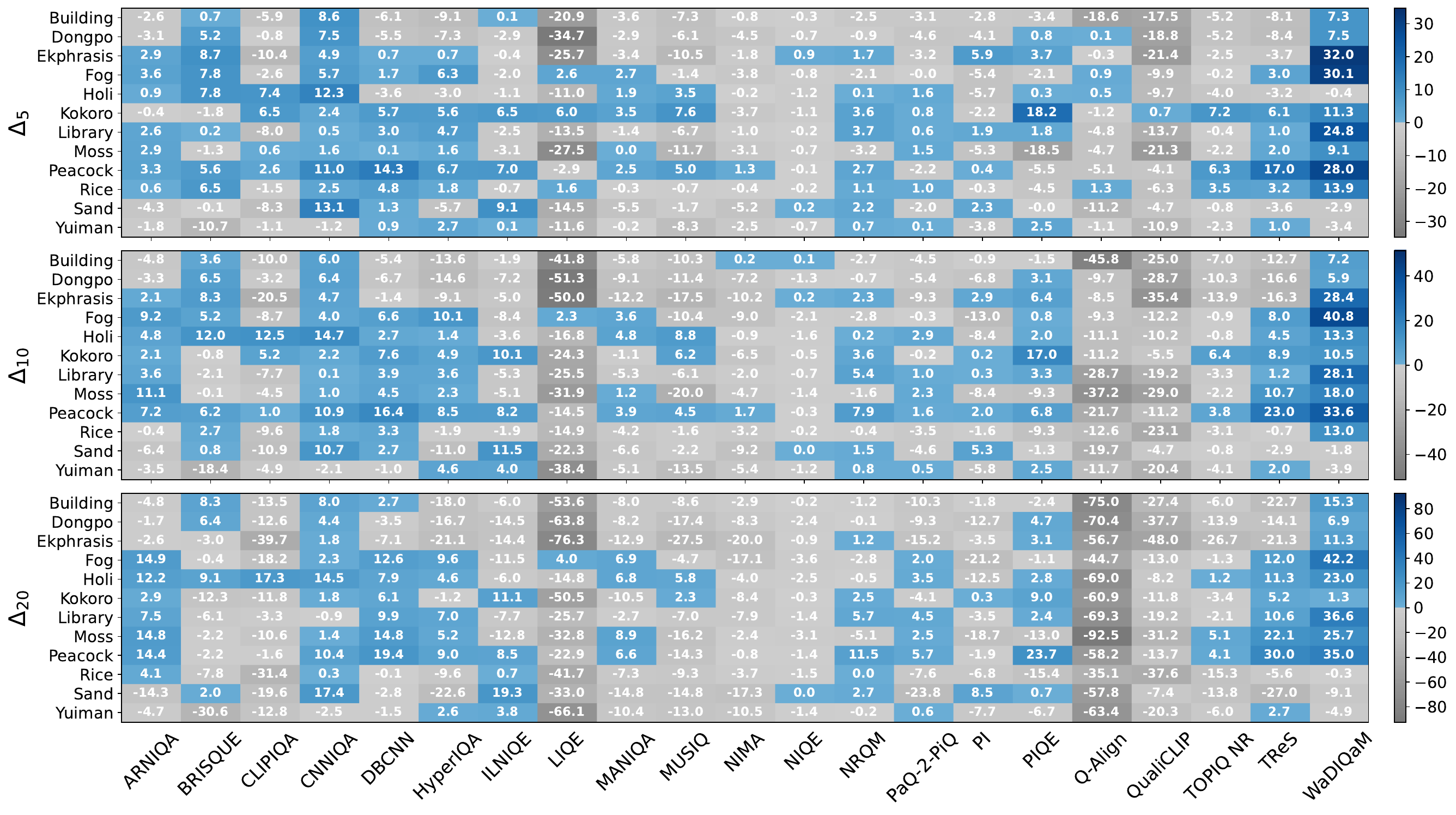} 
  \caption{\textbf{All 21 NR-IQA metrics fail on identifying degradation, assigning higher normalized scores to images with worse quality.} The heatmap shows the gap between the pair of normalized scores calculated for the image of a later editing (5, 10, or 20) and the image of the first step. The normalization converts each NR-IQA metric to the same scale of [0, 100], with higher scores corresponding to better image quality. Positive gaps indicate failures and are marked by blue colors in the heatmap. Due to the diversity in the texture types of the 12 initial images, each NR-IQA metric fails on a different set of initial images.}
  \label{fig:iqa_nano_banana_pro}
\end{figure}

\begin{figure}
  \centering
  \includegraphics[width=0.8\columnwidth]{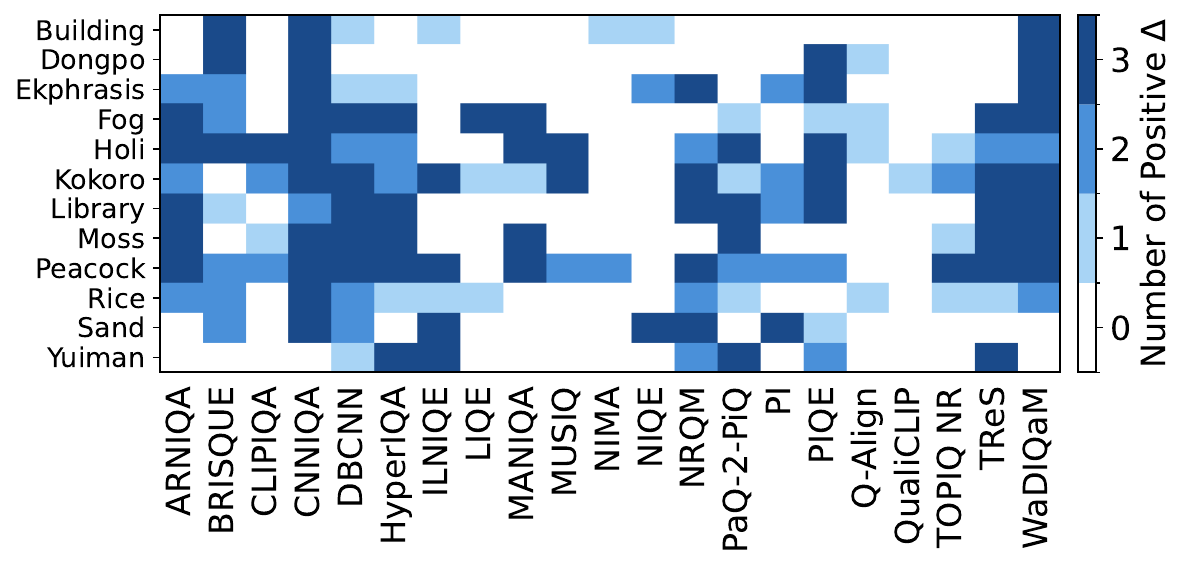} 
  \caption{\textbf{Aggregated results show that none of the 21 NR-IQA metrics fully succeed on all images.} This heatmap overlays the 3 heatmaps from \Cref{fig:iqa_nano_banana_pro}, and the brightness of the blue colors correspond to the total number of failures. None of the metrics show a fully white column, corresponding to consistent success.}
  \label{fig:iqa_nano_banana_pro_aggregated}
\end{figure}

Under this definition, a fully successful metric should have all three normalized score gaps ($\Delta_5$, $\Delta_{10}$, and $\Delta_{20}$) to be \textbf{negative}. The negative gaps indicate that a metric correctly identifies the image quality as degraded after 4, 9, and 19 steps. However, none of the 21 metrics (which are not based on large VLMs) fully succeeded (Figures \ref{fig:iqa_nano_banana_pro} and \ref{fig:iqa_nano_banana_pro_aggregated}). This suggests that the model-induced noise patterns confound these NR-IQA metrics. This could be explained by that these metrics are trained primarily on datasets constructed with heuristic distortions, such as KonIQ-10k~\cite{hosu2020koniq}, which qualitatively differ from the model-induced noise.

\subsection{Two Recent NR-IQA Methods Succeed}
\label{sec:two_recent_nr_iqa_methods_succeed}

\begin{figure}
  \centering
  \includegraphics[width=0.8\columnwidth]{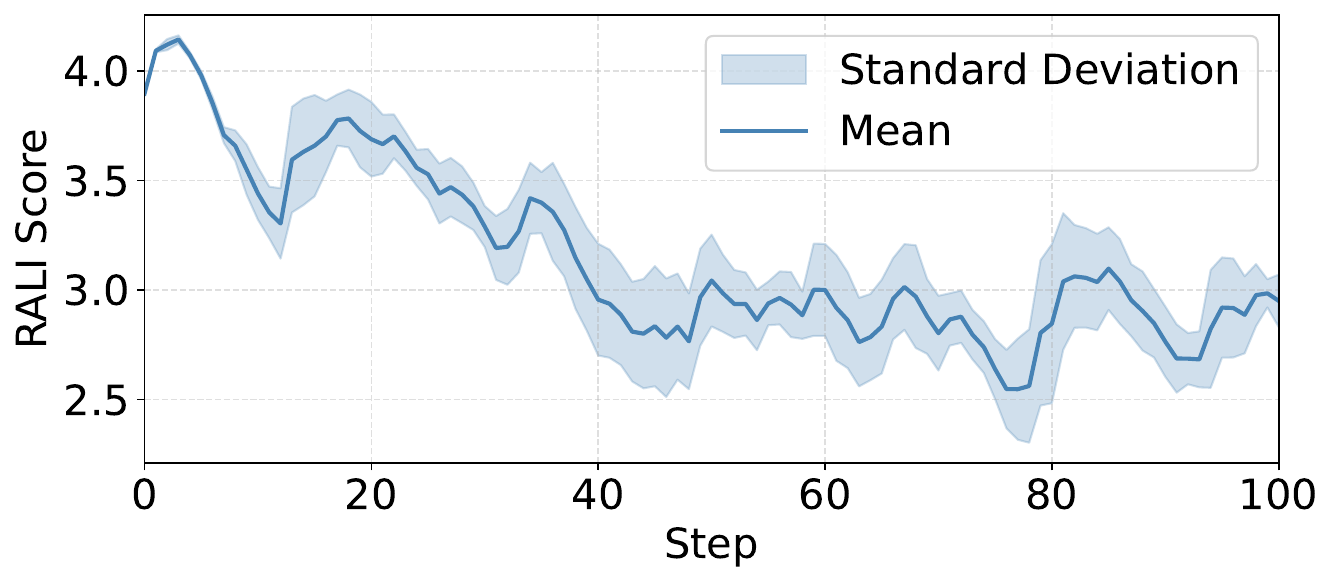} 
  \caption{\textbf{Despite a consistent drop in image quality, the RALI score (higher is better) fluctuates over the steps.} The fluctuation shows that RALI is not robust against the semantic change caused by iterative object addition (add-100-fruits).}
  \label{fig:rali_spikes}
\end{figure}

However, we highlight that RALI~\cite{zhao2025reasoning} and VisualQuality-R1~\cite{wu2025visualqualityr}, two recent large-VLM-based metrics, succeed on this task with 0 failure cases, although not free from other failure patterns. RALI is not robust against the change in the image content, exemplified by multiple spikes in the add-100-fruits run (\Cref{fig:rali_spikes}). VisualQuality-R1 had scores falling below 1, violating the lower-bound specified in its prompt. Despite these minor issues, the two recent NR-IQA methods successfully identify the accumulation of noise. The success of VisualQuality-R1 might be attributed to its training data covering a diverse mixture of IQA datasets.

\subsection{Self-Evaluation is Delayed}
\label{sec:self_evaluation_is_delayed}

In the reasoning summary, Nano Banana Pro comments on the original image in the generation section. The comment sometimes mentions the degradation, which can potentially serve a proxy to identify whether the generator is aware of the quality issue, circumventing the evaluator failures. 

To check whether the model comments on the noise, we use LLM-as-a-judge with Gemini-3-flash (prompt shown in \Cref{fig:self_evaluation_noise_prompt}). Out of the 100 steps, we looked for the first step where the answer is ``yes'', reporting average and standard deviation calculated over 5 replication runs. For the 12 initial images, the smallest identification step is 20 $\pm$ 4 for Holi, and the largest identification step is 37 $\pm$ 8 for Rice. These numbers are large compared to the step number where the introduced noise is very obvious, around 5 to 10. This suggests that the generator is not sensitive to the noise it generates, despite the reasoning summary exhibiting a certain extent of (heavily hallucinated) self-evaluation.

\begin{figure}[h]
\centering
\begin{promptbox}[title=LLM-as-a-Judge Prompt Template]
Below is a reasoning summary from an image editing model. Please identify if the reasoning summary mentions that the original image is noisy, pixelated, or contains visible artifacts. Output ``yes" or ``no" only.

\{reasoning\_summary\}
\end{promptbox}
\caption{\textbf{The LLM-as-a-judge prompt template to identify whether Nano Banana Pro acknowledges the noise during generation.} The reasoning summary, such as one shown in \Cref{fig:example_reasoning_summary}, will be inserted to the end of this prompt template.}
\label{fig:self_evaluation_noise_prompt}
\end{figure}

\begin{figure}
  \centering
  \includegraphics[width=1\columnwidth]{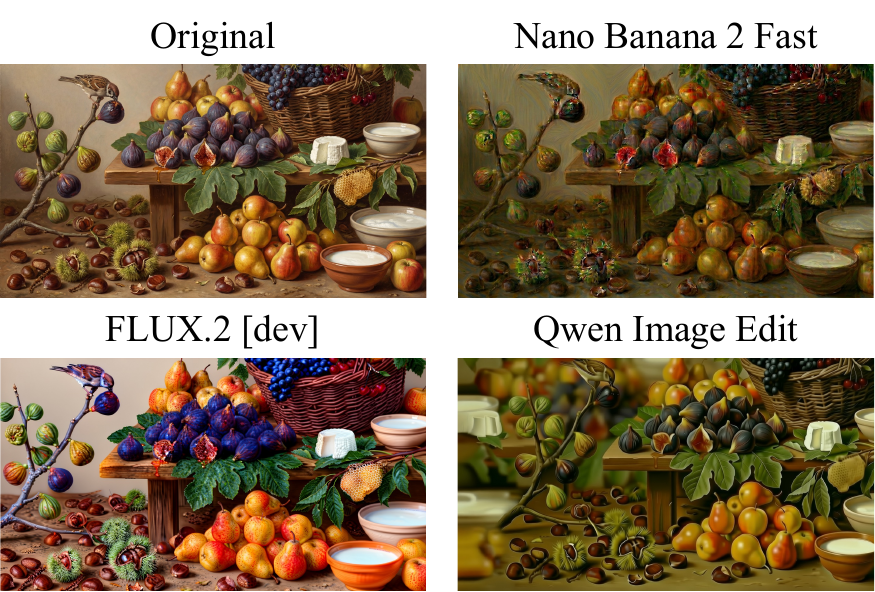}
  \caption{\textbf{Different noise accumulates during image replication by 3 more models.} Nano Banana 2 Fast (without reasoning) generates wrinkles that align with the contours of the objects. FLUX.2 [dev] generates scatter points on many of the objects. Qwen Image Edit simplifies the texture and erroneously duplicates objects on the right side of the image to the left side of the image.}
  \label{fig:other_models}
\end{figure}

\subsection{Other Image-Editing Models Fail Similarly}
\label{sec:other_image_editing_models_fail_similarly}

To examine if noise accumulation is pervasive across models, we follow the image generation and evaluation protocols using three alternative models: Nano Banana 2 Fast (without reasoning)~\cite{google2026nanobanana2}, FLUX.2 [dev]~\cite{bfl2025flux2}, and Qwen Image Edit~\cite{qwen2025imageedit2511, wu2025qwen}. We used these models to replicate each of the 12 seed images for 20 steps, repeated for 5 runs. We also used these models for two object addition runs. Overall, 1,400 new images were created using each model.

From the results, we saw that noise similarly accumulated over editing steps for each of the models we examined. Notably, the open-source models FLUX.2 [dev] and Qwen Image Edit also suffered from noise, suggesting that the watermarks in the proprietary Nano Banana model family~\cite{gowal2025synthid, googledeepmind2026synthid} are not the single cause for quality degradation. 

\begin{figure}
  \centering
  \includegraphics[width=0.8\columnwidth]{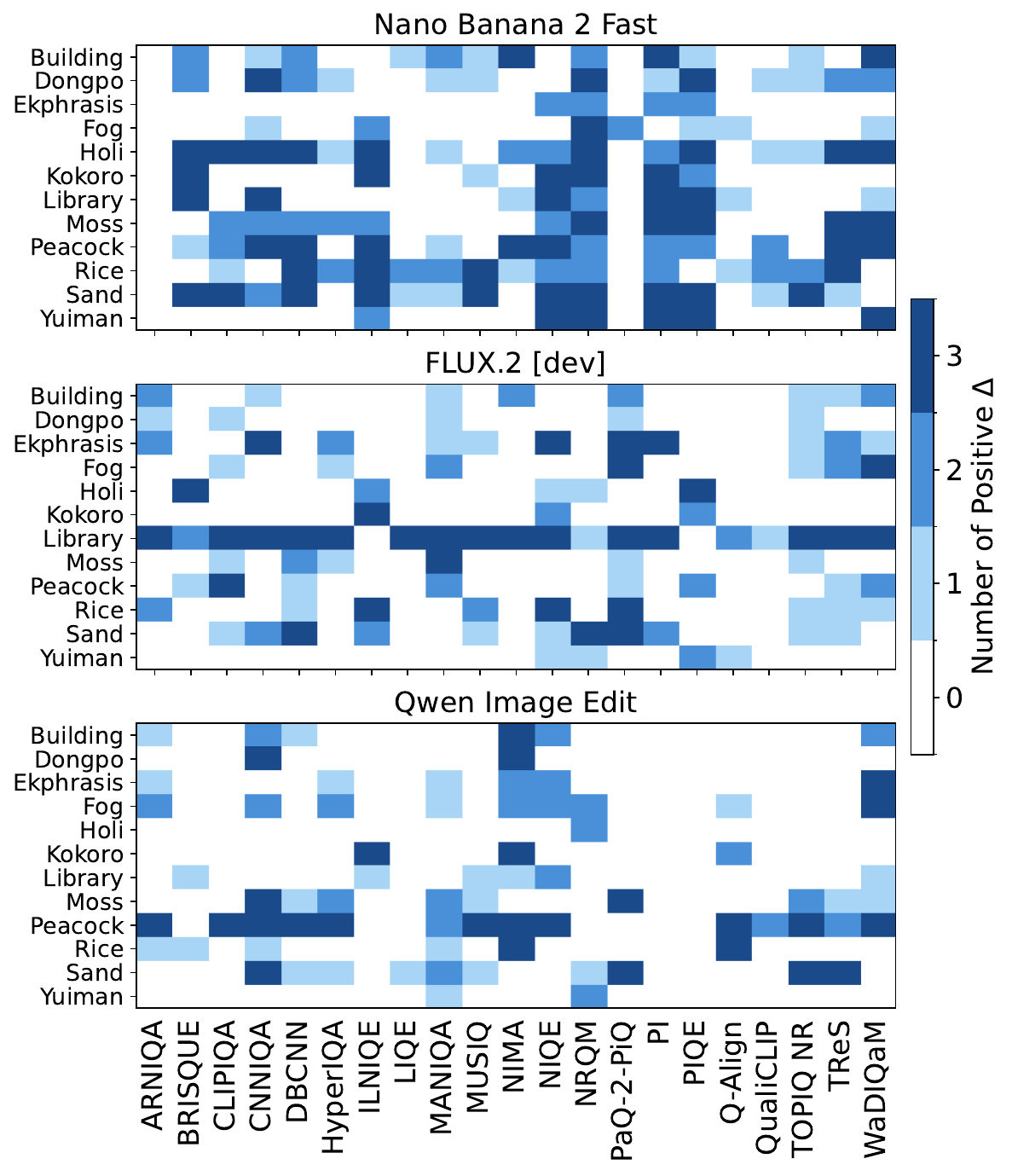} 
  \caption{\textbf{Similar to the evaluation of Nano-Banana-Pro generated results, NR-IQA metrics also fail for results from 3 more models.} No metric succeeds on all initial images and all models. Interestingly, PI and PIQE fully succeed on Qwen Image Edit, but fails for almost all initial images for Nano Banana 2 Fast. The diverse failure patterns across metrics further confirm the difference of the noise patterns from each model (\Cref{fig:other_models}).}
  \label{fig:iqa_other_models}
  \vspace{-0.5em}
\end{figure}

However, the noise accumulation patterns differ between these models (\Cref{fig:other_models}). A further test using 21 NR-IQA metrics reveal that the metrics again failed on these models, with different failure patterns confirming the qualitatively different nature of the noise patterns (\Cref{fig:iqa_other_models}). Due to the significant time investment, we did not run the most promising but very large VisualQuality-R1 model on these images.

\section{Conclusion}

Banana100 highlights the fragility of current image generators and evaluators in long-term image editing. By releasing 28,000 images that demonstrate quality degradation, we aim to facilitate the development of robust IQA metrics and degradation-free image editors, preventing the unintentional but unchecked pollution of the digital visual ecosystem.

{
    \small
    \bibliographystyle{ieeenat_fullname}
    \bibliography{main}
}


\end{document}